\documentclass[11pt,a4paper]{article}
\usepackage[hyperref]{acl2018}
\usepackage{times}
\usepackage{latexsym}
\usepackage{mdwlist}
\usepackage{tikz}
\usetikzlibrary{chains}
\usepackage{makecell}
\usepackage{url}
\usepackage{mdwlist}
\usepackage{fixltx2e}
\usepackage{xcolor}

\aclfinalcopy 
\def\aclpaperid{1537} 

\title{SciDTB: Discourse Dependency TreeBank for Scientific Abstracts}

\author{An Yang \qquad
	Sujian Li\footnote{Corresponding Author}  \\
	Key Laboratory of Computational Linguistics, Peking University, MOE, China \\
	{\tt \{yangan, lisujian\}@pku.edu.cn} \\}

\date{}

\begin{document}
\maketitle
\begin{abstract}
	
  Annotation corpus for discourse relations benefits NLP tasks such as machine translation and question answering. In this paper, we present SciDTB, a domain-specific discourse treebank annotated on scientific articles. 
  Different from widely-used RST-DT and PDTB, SciDTB uses dependency trees to represent discourse structure, which is flexible and simplified to some extent but do not sacrifice structural integrity. We discuss the labeling framework, annotation workflow and some statistics about SciDTB. Furthermore, our treebank is made as a benchmark for evaluating discourse dependency parsers, on which we provide several baselines as fundamental work.
  
\end{abstract}

\section{Introduction}

Discourse relation depicts how the text spans in a text relate to each other.
These relations can be categorized into different types according to semantics, logic or writer's intention. 
Annotations of such discourse relations can benefit many down-stream NLP tasks including machine translation \cite{guzman2014using, joty2014discotk} and automatic summarization \cite{gerani2014abstractive}.

Several discourse corpora have been proposed in previous work, grounded with various discourse theories. Among them Rhetorical Structure Theory TreeBank (RST-DT) \cite{carlson2003building} and Penn Discourse TreeBank (PDTB) \cite{prasad2007penn} are the most widely-used resources. 
PDTB focuses on shallow discourse relations between two arguments and ignores the whole organization.
RST-DT based on Rhetorical Structure Theory (RST) \cite{mann1988rhetorical} represents a text into a hierarchical discourse tree.
Though RST-DT provides more comprehensive discourse information, its limitations including the introduction of intermediate nodes and absence of non-projective structures bring the annotation and parsing complexity.

\newcite{li2014text} and \newcite{yoshida2014dependency} both realized the problems of RST-DT and introduced dependency structures into discourse representation. 
\newcite{stede2016parallel} adopted dependency tree format to compare RST structure and Segmented Discourse Representation Theory(SDRT) \cite{lascarides2008segmented} structure for a corpus of short texts.
Their discourse dependency framework is adapted from syntactic dependency structure \cite{hudson1984word,bohmova2003prague}, with words replaced by elementary discourse units (EDUs).
Binary discourse relations are represented from dominant EDU (called ``head'') to subordinate EDU (called ``dependent''), which makes non-projective structure possible.
However, \newcite{li2014text} and \newcite{yoshida2014dependency} mainly focused on the definition of discourse dependency structure and directly transformed constituency trees in RST-DT into dependency trees.
On the one hand, they only simply treated the transformation ambiguity, while constituency structures and dependency structures did not correspond one-to-one.
On the other hand, the transformed corpus still did not contain  non-projective dependency trees, though ``crossed dependencies'' actually exist in the real flexible discourse structures \cite{wolf2005representing}.
In such case, it is essential to construct a discourse dependency treebank from scratch instead of through automatically converting from the constituency structures.

In this paper, we construct the discourse dependency corpus SciDTB\footnote{The treebank is available at \href{https://github.com/PKU-TANGENT}{https://github.com/PKU-TANGENT/SciDTB}}. based on scientific abstracts, with the reference to the discourse dependency representation in \newcite{li2014text}.
We choose scientific abstracts as the corpus for two reasons.
First, we observe that when long news articles in RST-DT have several paragraphs, the discourse relations between paragraphs are very loose and their annotations are not so meaningful.
Thus, short texts with obvious logics become our preference. 
Here, we choose scientific abstracts from ACL Anthology\footnote{\href{http://www.aclweb.org/anthology/}{http://www.aclweb.org/anthology/}. SciDTB follows the same \href{https://creativecommons.org/licenses/by-nc-sa/3.0/deed.en}{CC BY-NC-SA 3.0} and \href{https://creativecommons.org/licenses/by/4.0/deed.en}{CC BY 4.0} licenses as ACL Anthology.} which are usually composed of one passage and have strong logics.
Second, we prefer to conduct domain-specific discourse annotation. RST-DT and PDTB are both constructed on news articles, which are unspecific in domain coverage.
We choose the scientific domain that is more specific and can benefit further academic applications such as automatic summarization and translation.
Furthermore, our treebank SciDTB can be made as a benchmark for evaluating discourse parsers. 
Three baselines are provided as fundamental work.

\section{Annotation Framework}

In this section, we describe two key aspects of our annotation framework, including elementary discourse units (EDU) and discourse relations.

\subsection{Elementary Discourse Units}

We first need to divide a passage into non-overlapping text spans, which are named elementary discourse units (EDUs).  
We follow the criterion of \citeauthor{polanyi1988formal} \shortcite{polanyi1988formal} and \citeauthor{irmer2011bridging} \shortcite{irmer2011bridging} which treats clauses as EDUs. 

However, since a discourse unit is a semantic concept but a clause is defined syntactically, in some cases segmentation by clauses is still not the most proper strategy. 
In practice, we refer to the guidelines defined by \cite{carlson2001discourse}. For example,
subjective clauses, objective clauses of non-attributive  verbs  and verb complement clauses are not segmented. 
Nominal postmodifiers with predicates are treated as EDUs. 
Strong discourse cues such as  ``\textit{despite}'' and ``\textit{because of}'' starts a new EDU no matter they are followed by a clause or a phrase.  
We give an EDU segmentation example as follows.

\begin{enumerate}
	\setlength{\itemsep}{0pt}
	\setlength{\partopsep}{-11pt}
	\item[1.] \textit{[Despite bilingual embedding’s success,][\textbf{the contextual information}][which is of critical importance to translation quality,][\textbf{was ignored in previous work.}]}
\end{enumerate}

It is noted,  as in Example 1, there are EDUs which are broken into two parts (in bold) by relative clauses or nominal postmodifiers. 
Like RST, we connect the two parts by a pseudo-relation type \textit{Same-unit} to represent their integrity.

\begin{table}[t!]
	\centering 
	\small
	\begin{tabular}{c|p{2cm}|p{4.5cm}}
		\hline\hline & \bf Coarse & \bf Fine \\ \hline\hline
		1.& ROOT & ROOT \\ \hline
		2.& Attribution & Attribution \\ \hline
		3.& Background &  Related, Goal, General \\ \hline
		4.& Cause-effect &  Cause, Result  \\ \hline
		5.& Comparison & Comparison  \\ \hline
		6.& Condition & Condition  \\ \hline
		7.& Contrast & Contrast \\ \hline
		8.& Elaboration & Addition, Aspect, Process-step, Definition, Enumerate, Example  \\ \hline
		9.& Enablement &  Enablement  \\ \hline
		10.& Evaluation & Evaluation  \\ \hline
		11.& Explain & Evidence, Reason \\ \hline
		12.& Joint & Joint  \\ \hline
		13.& Manner-means &  Manner-means \\ \hline
		14.& Progression & Progression  \\ \hline
		15.& Same-unit &  Same-unit \\ \hline
		16.& Summary & Summary \\ \hline
		17.& Temporal & Temporal \\ \hline\hline
	\end{tabular}
	\caption{\label{category}Discourse relation category of SciDTB. }
\end{table}

\subsection{Discourse Relations}

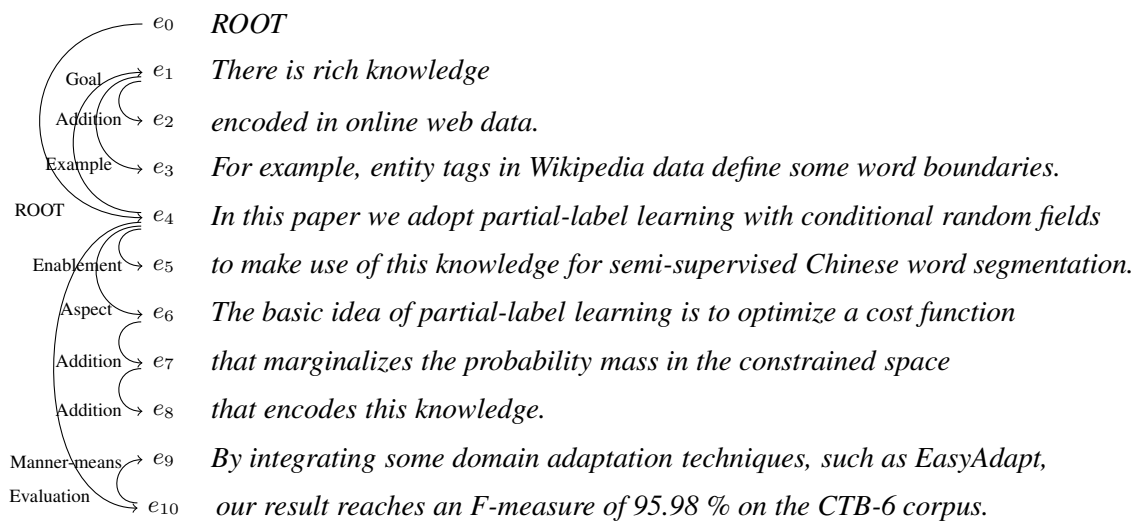
\begin{figure*}[ht]
	\centering
	\begin{tikzpicture}
	[node distance=2mm,
	content/.style={}]
	\node 	(e0)	{\small $e_0$};
	\node 	(e1)	[below=of e0]	{\small $e_1$};
	\node 	(e2)	[below=of e1]	{\small $e_2$};
	\node 	(e3)	[below=of e2]	{\small $e_3$};
	\node 	(e4)	[below=of e3]	{\small $e_4$};
	\node 	(e5)	[below=of e4]	{\small $e_5$};
	\node 	(e6)	[below=of e5]	{\small $e_6$};
	\node 	(e7)	[below=of e6]	{\small $e_7$};
	\node 	(e8)	[below=of e7]	{\small $e_8$};
	\node 	(e9)	[below=of e8]	{\small $e_9$};
	\node 	(e10)	[below=of e9]	{\small $e_{10}$};
	\node[content]   (t0)    [right=of e0]   {\textit{ROOT}};
	\node[content]   (t1)    [right=of e1]   {\textit{There is rich knowledge}};
	\node[content]   (t2)    [right=of e2]   {\textit{encoded in online web data.}};
	\node[content]   (t3)    [right=of e3]   {\textit{For example, entity tags in Wikipedia data define some word boundaries.}};
	\node[content]   (t4)    [right=of e4]   {\textit{In this paper we adopt partial-label learning with conditional random fields}};
	\node[content]   (t5)    [right=of e5]   {\textit{to make use of this knowledge for semi-supervised Chinese word segmentation.}};
	\node[content]   (t6)    [right=of e6]   {\textit{The basic idea of partial-label learning is to optimize a cost function}};
	\node[content]   (t7)    [right=of e7]   {\textit{that marginalizes the probability mass in the constrained space}};
	\node[content]   (t8)    [right=of e8]   {\textit{that encodes this knowledge.}};
	\node[content]   (t9)    [right=of e9]   {\textit{By integrating some domain adaptation techniques, such as EasyAdapt,}};
	\node[content]   (t10)    [right=of e10]   {\textit{our result reaches an F-measure of 95.98 \% on the CTB-6 corpus.}};
	\draw[->] (e0.west) .. controls +(180:1.8cm) and +(180:1.8cm).. (e4.west) node [very near end, left=0.3cm] {{\scriptsize ROOT}};
	\draw[->] (-0.3, -2.5) .. controls +(180:1.2cm) and +(180:1.2cm).. (e1.west) node [very near end, left] {{\scriptsize Goal}};
	\draw[->] (-0.3, -0.76) .. controls +(180:0.4cm) and +(180:0.4cm).. (e2.west) node [very near end, left] {{\scriptsize Addition}};
	\draw[->] (-0.3, -0.7) .. controls +(180:0.8cm) and +(180:0.8cm).. (e3.west) node [very near end, left] {{\scriptsize Example}};
	\draw[->] (-0.3, -2.72) .. controls +(180:0.4cm) and +(180:0.4cm).. (e5.west) node [very near end, left] {{\scriptsize Enablement}};
	\draw[->] (-0.3, -2.68) .. controls +(180:0.8cm) and +(180:0.8cm).. (e6.west) node [very near end, left] {{\scriptsize Aspect}};
	\draw[->] (-0.3, -3.95) .. controls +(180:0.4cm) and +(180:0.4cm).. (e7.west) node [very near end, left] {{\scriptsize Addition}};
	\draw[->] (-0.3, -4.57) .. controls +(180:0.4cm) and +(180:0.4cm).. (e8.west) node [very near end, left] {{\scriptsize Addition}};
	\draw[->] (-0.3, -2.63) .. controls +(180:1.5cm) and +(180:1.5cm).. (e10.west) node [very near end, left] {{\scriptsize Evaluation}};
	\draw[->] (-0.35,-6.35) .. controls +(180:0.4cm) and +(180:0.4cm).. (e9.west) node [very near end, left] {{\scriptsize Manner-means}};
	\end{tikzpicture}		
	\caption{\label{anno_example}An example discourse dependency tree for an abstract in SciDTB.}
\end{figure*}

A discourse relation is defined as tri-tuple $(h, d, r)$, where $h$ means the head EDU, $d$ is the dependent EDU, and $r$ defines the relation category between $h$ and $d$.
For a discourse relation, head EDU is defined as the unit with essential information and dependent EDU  with supportive content.
	Here, we follow \newcite{carlson2001discourse} to adopt deletion test in the determination of head and dependent. 
	If one unit in a binary relation pair is deleted but the whole meaning can still be almost understood from the other unit, the deleted unit is treated as dependent and the other one as the head.

For the relation categories, we mainly refer to the work of \cite{carlson2001discourse} and \cite{bunt2016iso}. 
Table~\ref{category} presents the discourse relation set of SciDTB, which are not explained detailedly one by one due to space limitation.
Through investigation of scientific abstracts, we define 17 coarse-grained relation types and 26 fine-grained relations for SciDTB. 

It is noted that we make some modifications to adapt to the scientific domain.
For example, In SciDTB, \textit{Background} relation is divided into three subtypes: \textit{Related}, \textit{Goal} and \textit{General}, because the background description in scientific abstracts usually has more different intents.
Meanwhile, for \textit{attribution} relation we treat the attributive content rather than act as head, which is contrary to that defined in \cite{carlson2001discourse}, because scientific facts or research arguments mentioned in attributive content are more important in abstracts. 
For some symmetric discourse relations such as \textit{joint} and \textit{comparison}, 
where two connected EDUs are equally important and have interchangeable semantic roles, we follow the strategy as \cite{li2014text} and treat the preceding EDU as the head.

Another issue on coherence relations is about polynary relations which involve more than two EDUs. 
	The first scenario is that one EDU dominates a set of posterior EDUs as its member.
	In this case, we annotate binary relations from head EDU to each member EDU with the same relation. 
	The second scenario is that several EDUs are of equal importance in a polynary relation. 
	For this case, we link each former EDU to its neighboring EDU with the same relation, forming a relation chain similar to ``right-heavy'' binarization transformation in \cite{morey2017much}.

By assuring that each EDU has one and only one head EDU, we can obtain a dependency tree for each scientific abstract. An example of dependency annotation is shown in Figure~\ref{anno_example}.

\section{Corpus Construction}

Following the annotation framework, we collected 798 abstracts from ACL anthology and constructed the SciDTB corpus. The construction details are introduced as follows.

\paragraph{Annotator Recruitment} 

To select annotators, we put forward two requirements to ensure the annotation quality. First, we required the candidates to have linguistic knowledge. 
Second, each candidate was asked to join a test annotation of 20 abstracts, whose quality was evaluated by experts.
After the judgement, 5 annotators were qualified to participate in our work.

\paragraph{EDU Segmentation}

We performed EDU segmentation in a semi-automatic way. 
First, we did sentence tokenization on raw texts using NLTK 3.2 \cite{bird2004nltk}. Then we used SPADE \cite{soricut2003sentence}, a pre-trained EDU segmenter relying on Charniak's syntactic parser \cite{charniak2000maximum}, to automatically cut sentences into EDUs. 
Then, we manually checked each segmented abstract to ensure the segmentation quality.
Two annotators conducted the checking task, with one proofreading the output of SPADE, and the other reviewing the proofreading. 
The checking process was recorded for statistical analysis.

\paragraph{Tree Annotation}
Labeling dependency trees was the most labor-intensive work in the corpus construction. 
798 segmented abstracts were labeled by 5 annotators in 6 months. 
506 abstracts were annotated more than twice separately by different annotators, with the purpose of analysing annotation consistency and providing human performance as an upper bound.  
The annotated trees  were stored in JSON format.
For convenience, we developed an online tool\footnote{\href{http://123.56.88.210/demo/depannotate/}{http://123.56.88.210/demo/depannotate/}} for annotating and visualising discourse dependency trees. 

\section{Corpus Statistics}
SciDTB contains 798 unique abstracts with 63\% labeled more than once and 18,978 discourse relations in total.
Table~\ref{size} compares the size of SciDTB with RST-DT and another PDTB-style domain-specific corpus BioDRB \cite{prasad2011biomedical},
we can see SciDTB has a comparable size  with RST-DT. 
Moreover, it is relatively easy for SciDTB to augment its size since the dependency structure simplifies the annotation to some extent. 
Compared with BioDRB,  SciDTB has larger size and passage-level representations.
\begin{table}[ht]
	\small
	\begin{center}
		\begin{tabular}{c|c|c|c}
			\hline \textbf{Corpus} & \textbf{\#Doc.} & \textbf{\#Doc. (unique)} &  \textbf{\#Relation} \\ \hline
			\textbf{SciDTB} & 1355 & 798 & 18978 \\
			\textbf{RST-DT} & 438 & 385 & 23611 \\
			\textbf{BioDRB} & 24 & 24 & 5859\\
			\hline
		\end{tabular}
	\end{center}
	\caption{\label{size}Size of SciDTB and other discourse relation banks. }
\end{table}

\subsection{Annotation Consistency}

\paragraph{EDU Segmentation} 

We use 214 abstracts for analysis. After the proofreading of the first annotator, the abstracts are composed of totally 2,772 EDUs. 
Among these EDUs, only 28 (1.01\%) EDUs are disagreed and revised by the second annotator, which means  a very  high consensus between annotators on EDU segmentation.

\paragraph{Tree Labeling} 
\begin{table}[ht]
	\small
	\begin{center}
		\begin{tabular}{c|c|c|c|c}
			\hline \textbf{Annotator} & \textbf{\#Doc.} & \textbf{UAS} & \textbf{LAS} &  \textbf{Kappa score}\\ \hline
			{ 1 \&  2} & 93 & 0.811 & 0.644 & 0.763\\
			{ 1 \&  3} & 147 & 0.800 & 0.628  & 0.761\\
			{ 1 \&  4} & 42 & 0.772 & 0.609  & 0.767\\
			{ 3 \&  4} & 46 & 0.806 & 0.639  & 0.772\\
			{ 4 \&  5} & 44 & 0.753 & 0.550  & 0.699\\
			\hline
		\end{tabular}
	\end{center}
	\caption{\label{consistency}Relation annotation consistency. }
\end{table}

Here, we evaluate the consistency of two annotators on labeling discourse relations using 3 metrics from different aspects.
When labeling a discourse relation, each non-root EDU must choose its head with a specific relation type. 
Thus, the annotation disagreement mainly comes from selecting head or determining relation type.
Similar to syntactic dependency parsing, unlabeled and labeled attachment scores (\textbf{UAS} and \textbf{LAS}) are employed to measure the labeling correspondence.
UAS calculates the proportion of EDUs which are assigned the same head in two annotations, while LAS considers the uniformity of both head and relation label.
\textbf{Cohen's Kappa score} evaluates the agreement of labeling relation types under the premise of knowing the correct heads. 

Table~\ref{consistency} shows the agreement results between two annotators.
We can see that most of the \textbf{LAS} values between annotators exceed 0.60. The agreement on tree structure reflected by \textbf{UAS} all reaches 0.75. 
The \textbf{Kappa} values for relation types agreement keep equal to or greater than 0.7.

\subsection{Structural Characteristics}

\paragraph{Non-projection in Treebank} 

One advantage of dependency trees is that they can represent non-projective structures. 
In SciDTB, we annotated 39 non-projective dependency trees, which account for about 3\% of the whole corpus.
This phenomenon in our treebank is not so frequent as \cite{wolf2005representing}.
We think this may be because scientific abstracts are much shorter and scientific expressions are relatively restricted.

\paragraph{Dependency Distance}

\begin{table}[t]
	\small
	\begin{center}
		\begin{tabular}{c|c|c}
			\hline \textbf{Distance} & \textbf{\#Relations} & \textbf{Percentage/\%} \\ \hline
			{0 EDU} & 10576 & 61.64 \\
			{1 EDU} & 2208 & 12.87 \\
			{2 EDUs} & 1231 & 7.17 \\
			{3-5 EDUs} & 1626 & 9.48 \\
			{6-10 EDUs} & 1146 & 6.68 \\
			{11-15 EDUs} & 304 & 1.77 \\
			{$>$15 EDUs} & 67 & 0.39 \\ \hline
			{Total} & 17158 & 100.00 \\
			\hline
		\end{tabular}
	\end{center}
	\caption{\label{distrib}Distribution of dependency distance.}
\end{table}

Here we investigate the distance of two EDUs involved in a discourse relation.
The distance is defined as the number of EDUs between head and dependent. 
We present the distance distribution of all the relations in SciDTB, as shown in Table~\ref{distrib}. 
It should be noted that \textit{ROOT} and \textit{Same-unit} relations are omitted in this analysis.
From Table~\ref{distrib}, we find most relations connect near EDUs.
Most relations (61.6\%) occur between neighboring EDUs and about 75\% relations occur with at most one intermediate EDU.

Although most dependency relations function intra-sentence, there exist long-range dependency relations in the treebank.
On average, the distance of 8.8\% relations is greater than 5.
We summarize that the most frequent 5 fine-grained relation types of these long-distance relations belong to \textit{Evaluation}, \textit{Aspect}, \textit{Addition}, \textit{Process-step} and \textit{Goal}, which tend to appear on higher level in dependency trees.

\section{Benchmark for Discourse Parsers}

We further apply SciDTB as a benchmark for comparing and evaluating discourse dependency parsers. 
For the 798 unique abstracts in SciDTB, 154 are used for development set and 152 for test set. The remaining 492 abstracts are used for training. 
We implement two transition-based parsers and a graph-based parser as baselines.

\paragraph{Vanilla Transition-based Parser} 

We adopt the transition-based method for dependency parsing by \newcite{nivre2003efficient}.
The action set of arc-standard system \cite{nivre2004memory} is employed. 
We build an SVM classifier to predict most possible transition action for given configuration. 
We adopt the N-gram features, positional features, length features and dependency features for top-2 EDUs in the stack and top EDU in the buffer, which can be referred from \cite{li2014text,wang2017two}

\paragraph{Two-stage Transition-based Parser}

We implement a two-stage transition-based dependency parser following \cite{wang2017two}. First, an unlabeled tree is produced by vanilla transition-based approach. Then we train a separate SVM classifier to predict relation types on the tree in pre-order. For the 2nd-stage, apart from features in the 1st-stage, two kinds of features are added, including depth of head and dependent in the tree and the predicted relation between the head and its head.

\paragraph{Graph-based Parser} 
We implement a graph-based parser as in \cite{li2014text}. For simplicity, we use averaged perceptron rather than MIRA to train weights. N-gram, positional, length and dependency features between head and dependent labeled with relation type are considered.


\paragraph{Hyper-parameters}
During training, the hyper-parameters of these models are tuned using development set. 
	For vanilla transition-based parser, we take linear kernel for the SVM classifier. The penalty parameter C is set to 1.5. 
	For two-stage parser, the 1st-stage classifier follows the same setting as the vanilla parser. For 2nd-stage, we use the linear kernel and set C to 0.5.
	The averaged perceptron in graph-based parser is trained for 10 epochs on the training set. Weights of features are initialized to be 0 and trained with fixed learning rate.

\paragraph{Results}
Table~\ref{performance} shows the performance of these parsers on development and test data. 
We also measure parsing accuracy with UAS and LAS.
The human agreement is presented for comparison.
With the addition of tree structural features in relation type prediction, the two-stage dependency parser gets better performance on LAS than vanilla system on both development and test set.
Compared with graph-based model, the two transition-based baselines achieve higher accuracy with regard to UAS and LAS. 
Using more effective training strategies like MIRA may improve graph-based models. 
We can also see that human performance is still much higher than the three parsers, meaning there is large space for improvement in future work.
\begin{table}[t]
	\small
	\begin{center}
		\begin{tabular}{c|c|c|c|c}
			\hline & \multicolumn{2}{c|}{\textbf{Dev set}} & \multicolumn{2}{c}{\textbf{Test set}} \\
			\cline{2-5} & \textbf{UAS} & \textbf{LAS} & \textbf{UAS} & \textbf{LAS}\\ \hline
			Vanilla transition & \textbf{0.730} & 0.557 & \textbf{0.702} & 0.535\\
			Two-stage transition & \textbf{0.730} & \textbf{0.577} & \textbf{0.702} & \textbf{0.545}\\
			Graph-based & 0.607 & 0.455 & 0.576 & 0.425\\ \hline
			Human & 0.806 & 0.627 & 0.802 & 0.622 \\
			\hline
		\end{tabular}
	\end{center}
	\caption{\label{performance}Performance of baseline parsers. }
\end{table}

\section{Conclusions}

In this paper, we propose to construct a discourse dependency treebank  SciDTB for scientific abstracts. 
It represents passages with dependency tree structure, which is simpler and more flexible for analysis. 
We have presented our annotation framework, construction workflow and statistics of SciDTB, which can provide annotation experience for extending to other domains.
Moreover, this treebank can serve as an evaluating benchmark of discourse parsers.

 In the future, we will enlarge our annotation scale to cover more domains and longer passages, and 
 explore how to use SciDTB in some downstreaming applications. 
	

\section*{Acknowledgments}

We thank the anonymous reviewers for their insightful comments on this paper. 
We especially thank Liang Wang for developing the online annotation tool.
This work was partially supported by National Natural Science Foundation of China (61572049 and 61333018). 
The correspondence author of this paper is Sujian Li.

\bibliography{acl2018}
\bibliographystyle{acl_natbib}

\end{document}